\documentclass[runningheads]{llncs}
\usepackage[T1]{fontenc}
\usepackage{graphicx}
\usepackage{csquotes}
\usepackage{booktabs}
\usepackage{todonotes}
\usepackage{tcolorbox}
\usepackage[colorlinks=true, linkcolor=blue, citecolor=blue, urlcolor=blue, filecolor=blue, pdfborder={0 0 0}]{hyperref}
\usepackage{color}

\begin{document}

\title{Annotating Errors in English Learners' Written Language Production: Advancing Automated Written Feedback Systems}

\titlerunning{Annotating Learner Errors for Automated Feedback}

\author{Steven Coyne\inst{1, 2}\orcidID{0009-0008-2728-699X} \and
Diana Galvan-Sosa\inst{3}\orcidID{0000-0002-2997-6774} \and
Ryan Spring\inst{1}\orcidID{0000-0003-4810-9825} \and
Camélia Guerraoui\inst{1, 2, 4}\orcidID{0009-0009-1842-9941}\and
Michael Zock \inst{5}\orcidID{0000-0002-6169-0477} \and
Keisuke Sakaguchi\inst{1, 2}\orcidID{0000-0002-3809-1732} \and
Kentaro Inui\inst{6, 1, 2}\orcidID{0000-0001-6510-604X}}

\authorrunning{S. Coyne et al.}
\institute{Tohoku University, Sendai, Japan \and
RIKEN, Wako, Japan \and
ALTA Institute, Computer Laboratory, University of Cambridge, U.K \and
INSA Lyon, Villeurbanne, France \and
CNRS, LIS, Aix-Marseille University, Marseille, France \and
MBZUAI, Abu Dhabi, United Arab Emirates
\\
\email{coyne.steven.charles.q2@dc.tohoku.ac.jp}}

\maketitle

\begin{abstract}

Recent advances in natural language processing (NLP) have contributed to the development of automated writing evaluation (AWE) systems that can correct grammatical errors. 
However, while these systems are effective at improving text, they are not optimally designed for language learning.
They favor direct revisions, often with a click-to-fix functionality that can be applied without considering the reason for the correction.
Meanwhile, depending on the error type, learners may benefit most from simple explanations and strategically indirect hints, especially on generalizable grammatical rules.
To support the generation of such feedback, we introduce an annotation framework that models each error's error type and generalizability.
For error type classification, we introduce a typology focused on inferring learners' knowledge gaps by connecting their errors to specific grammatical patterns.
Following this framework, we collect a dataset of annotated learner errors and corresponding human-written feedback comments, each labeled as a direct correction or hint.
With this data, we evaluate keyword-guided, keyword-free, and template-guided methods of generating feedback using large language models (LLMs).
Human teachers examined each system's outputs, assessing them on grounds including relevance, factuality, and comprehensibility.
We report on the development of the dataset and the comparative performance of the systems investigated.

\end{abstract}

\section{Introduction}\label{sec:introduction}

Millions of people worldwide study English as a second language (L2), driven in part by its role in professional and academic domains.
Research in education and applied linguistics has shown that learning is enhanced by targeted, specific, and timely feedback \cite{goodwin2012timely}, with recent meta-analyses finding a beneficial impact of written corrective feedback (WCF) on L2 learners' language development \cite{Brown2023survey,Kang2015TheEO}.
However, writing high-quality feedback is a resource-intensive task for educators, limiting its availability.
Advances in natural language processing (NLP) offer promising solutions to this challenge, potentially enabling automated systems that can provide learners with consistent and accessible WCF.

However, the development of effective automated feedback systems requires careful consideration of pedagogical principles exemplified by human teachers' practices, which we examine in Section \ref{sec:principles_of_wcf}.
Specifically, we seek to annotate data to model 1) the teacher's interpretation of what knowledge gap an error represents, and 2) the practice of adapting feedback strategy to the details of the error.

Our research makes four contributions to the field of automated WCF: First, we develop an annotation framework that explicitly models both error characteristics and feedback strategies.
As detailed in Section \ref{sec:annotation_scheme}, we focus on the \textbf{error type}, its \textbf{generalizability} (``treatability''), and the \textbf{directness} of a feedback comment.
Second, we introduce an error typology that better aligns with educational perspectives than existing frameworks, enabling more targeted feedback generation.
Third, we gather a dataset of learner errors and associated feedback comments annotated with our framework, described in Section \ref{sec:annotating_data}.
Fourth, we empirically validate the effectiveness of our approach through experiments with LLM-based feedback generation using keyword-guided systems, a template-guided system, and a keyword-free system (detailed in Section \ref{sec:system_experiments}), followed by manual evaluation of the outputs by practicing English teachers (Section \ref{sec:rating_experiments}).

\begin{figure}[t]
    \centering
    \includegraphics[width=0.7\textwidth]{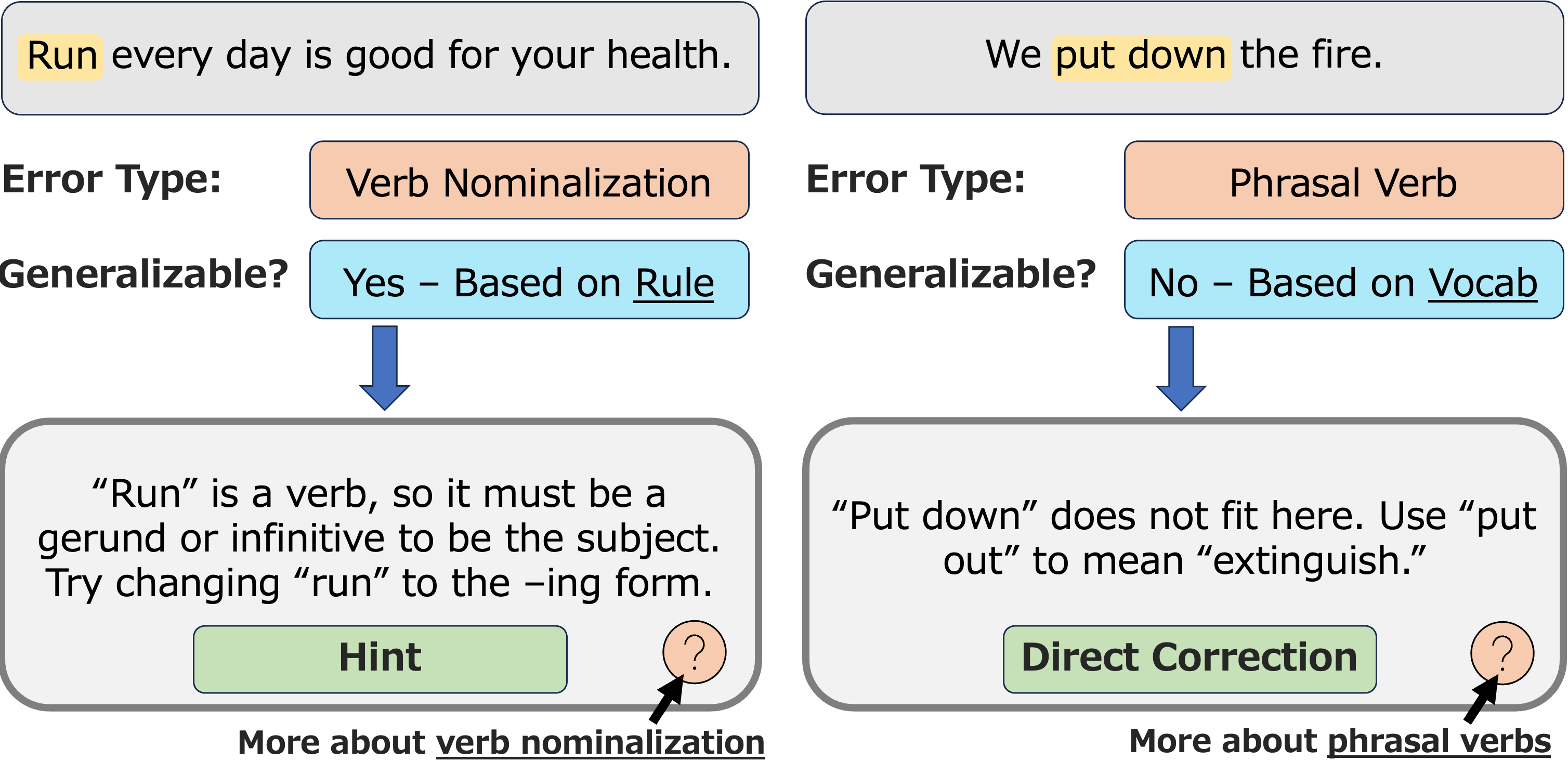}
    \caption{Example of feedback utilizing our framework, with feedback strategy differentiated based on error type and generalizability.}
    \label{fig:differentiated-feedback}
\end{figure}

We report our results in Section \ref{sec:results}, finding that all systems perform well in this task, with keyword systems and the keyword-free system performing similarly.
While the template system performed relatively well when appropriate templates were available, template coverage gaps resulted in mixed or incoherent feedback comments.
Furthermore, when no appropriate template option was available, the template system struggled to refrain from giving feedback, selecting suboptimal templates instead.
However, the template system had the highest level of agreement on directness with human feedback writers.

To support future research in automated feedback, we release our guidelines and resources, available at \href{https://github.com/coynestevencharles/annotating-errors-wcf}{github.com/coynestevencharles/annotating-errors-wcf}.

\section{Principles of Written Corrective Feedback}\label{sec:principles_of_wcf}

\begin{figure*}[!ht]
\centering\includegraphics[width=0.9\textwidth]{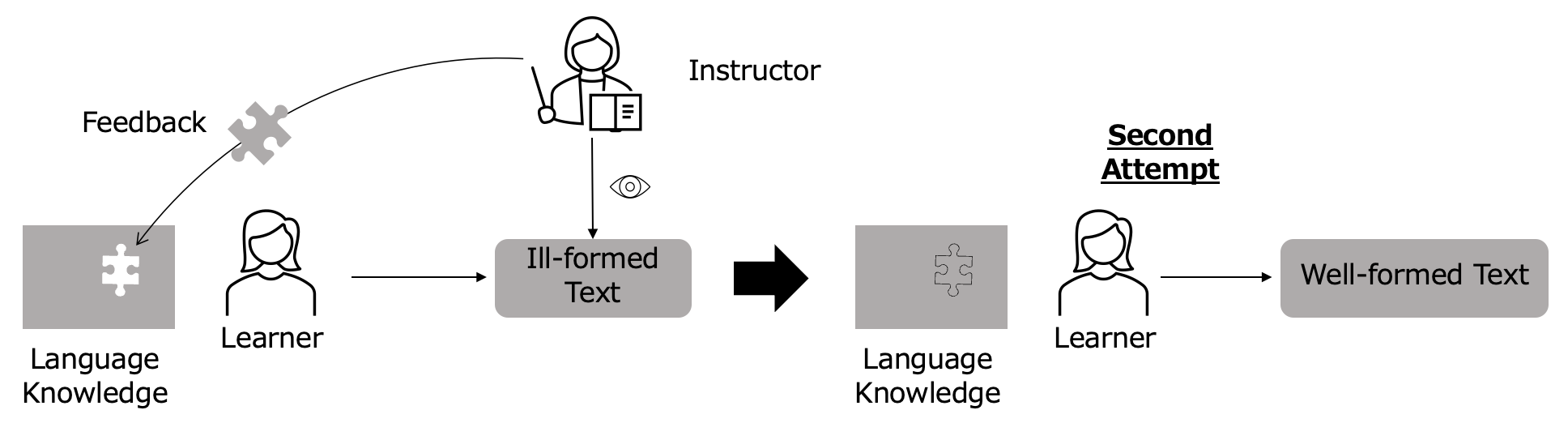}
    \caption{The cycle of feedback in a learning setting.}
    \label{fig:feedback-cycle}
\end{figure*}

Applied linguists have studied corrective feedback for decades.
Works such as Bitchener et al. \cite{Bitchener_2021}, Ferris \cite{ferris_treatment_2011}, Sheen \cite{sheen_corrective_2011}, and Ellis \cite{ellis-2008-typology} provide detailed overviews of the topic, describing a number of dimensions of feedback and the factors contributing to a given feedback comment.
In this section, we examine the nature of WCF as an act of communication within a cyclical feedback setting, and identify the factors that teachers use to determine the content and delivery strategy of their feedback comments.
Specifically, we seek to identify  ``key'' factors that can be annotated in learner writing corpora in a practical way.

\subsubsection{The Feedback Cycle}

Feedback in a language education setting is part of a cycle of communication between a learner and an instructor or peer.
At first, the learner has partial knowledge of a concept and attempts to apply it.
The result may be flawed, such as a sentence with an error.
Based on the error, an instructor infers a knowledge gap and intervenes with feedback to help the learner improve.
The learner processes the feedback and may make a second attempt.
They may be successful on the new attempt, or the cycle may continue as long as their output is monitored by the instructor.
Figure \ref{fig:feedback-cycle} illustrates this learning cycle.

Alignment with this feedback cycle is an important aspect of this work, as it assumes that importance is placed on the development of the learner's skills over potentially multiple attempts.
In particular, the assumption of a feedback cycle is one motivation to limit the explicitness of feedback and prefer hints that invoke reflection and self-correction.

\subsubsection{Strategies for Giving Feedback} \label{sec:strategies_for_feedback}

Teachers providing feedback make various decisions about content and delivery, including what errors to target, how explicit to be, and what form the feedback will take \cite{Bitchener_2021}.
In this work, we focus on metalinguistic feedback \cite{ellis-2008-typology}, which provides information about the causes of errors and how to fix them.
It can vary considerably in complexity, directness of an edit suggestion (if included), and the extent of additional information known as ``elaboration'' \cite{Kulhavy1989,shute_2008,narciss2013}.
In the context of this work, the most important distinction to make is the difference between direct corrections and hints.

Teacher decisions about feedback are influenced by many factors, which we divide into three general groups. \textit{Task and setting factors} include the educational context of the writing task (e.g., ESL vs. EFL) and assignment type (e.g., daily writing vs. academic publication). 
\textit{Learner factors} include age, language level, first language, and past patterns of errors.
Finally, there are \textit{error factors}, which depend on the textual product itself.
We focus on error factors in this work, as it is more practical to gather text than metadata about real-world context.

One key error factor is the \textbf{error type} of the learner's error.
As feedback should be targeted and specific to the error in question \cite{goodwin2012timely}, it is natural for feedback to differ between e.g., a tense error and a spelling error.

An additional key error factor is the \textbf{generalizability} of the error, which has also been referred to as \textit{treatability}.
Ferris \cite{FERRIS19991} describes treatability as whether ``there are rules to consult'' for an error, or whether the error  is ``non-idiomatic'' or ``idiosyncratic.''
Ferris, Hyland, and Hyland \cite{Ferris_Hyland_Hyland_2006} reported that teachers gave direct feedback to untreatable errors 65.3\% of the time, and more indirect feedback 33.6\% of the time, whereas for \textit{treatable} errors the feedback was 36.7\% direct and 58.7\% indirect.
This pattern reflects research suggesting that direct corrections are more suitable when the error is based on a non-generalizable, potentially lexical issue, whereas indirect feedback or elaborated hints are more effective for generalizable errors based on broader grammatical rules  \cite{lee_2013,Brown2023survey}.

\section{Annotating Data for Feedback Generation}\label{sec:annotating_data}

In this section, we introduce our framework for annotating English learner writing data to facilitate educational feedback comment generation tasks.
We first review previous error typologies and annotation schemes, highlighting the need for a new approach.
Then, we detail our annotation scheme, which captures both error characteristics and feedback strategies.

\subsection{Related Work} \label{sec:annotation_related_work}

When designing our annotation framework and error typology, we considered existing error tag sets in NLP, corpus linguistics, and applied linguistics.

There are several typologies of errors used in the field of grammatical error correction (GEC).
These include the system used in the NUCLE dataset \cite{dahlmeier-etal-2013-building}, the typology used in the Cambridge Learner Corpus \cite{Nicholls2003TheCL} and its associated GEC corpora such as the First Certificate in English (FCE) dataset \cite{yannakoudakis-etal-2011-new}, and ERRANT \cite{bryant-etal-2017-automatic}, currently the most widely applied typology of errors in GEC research. 

However, these typologies primarily model surface-level errors for the purpose of correction, without capturing the underlying language knowledge gaps necessary for effective feedback.
A tag like ``Missing Preposition'' indicates the surface error, but does not specify whether it involves a phrasal verb, a prepositional phrase, or another grammatical pattern.
This distinction is crucial for providing feedback addressing the specific concept the learner is struggling with.

Recent research in error explanation \cite{song-etal-2024-gee,kaneko-okazaki-2024-controlled-generation} focuses on generating technical descriptions of errors, but these explanations are not written with language learners in mind, and not all function as metalinguistic feedback comments.

Concerning error generalizability and feedback directness, we are not aware of any dataset featuring such annotations for language learners' errors.
Furthermore, we are not aware of any well-accepted, comprehensive list of errors and whether they are \textit{generalizable}/\textit{treatable}.
Presenting a dataset with explicit generalizability and feedback directness labels thus a contribution of this work.

\subsection{Annotation Scheme}\label{sec:annotation_scheme}

\subsubsection{Error Annotations}

\begin{figure}[!t]
    \centering\includegraphics[width=0.7\columnwidth]{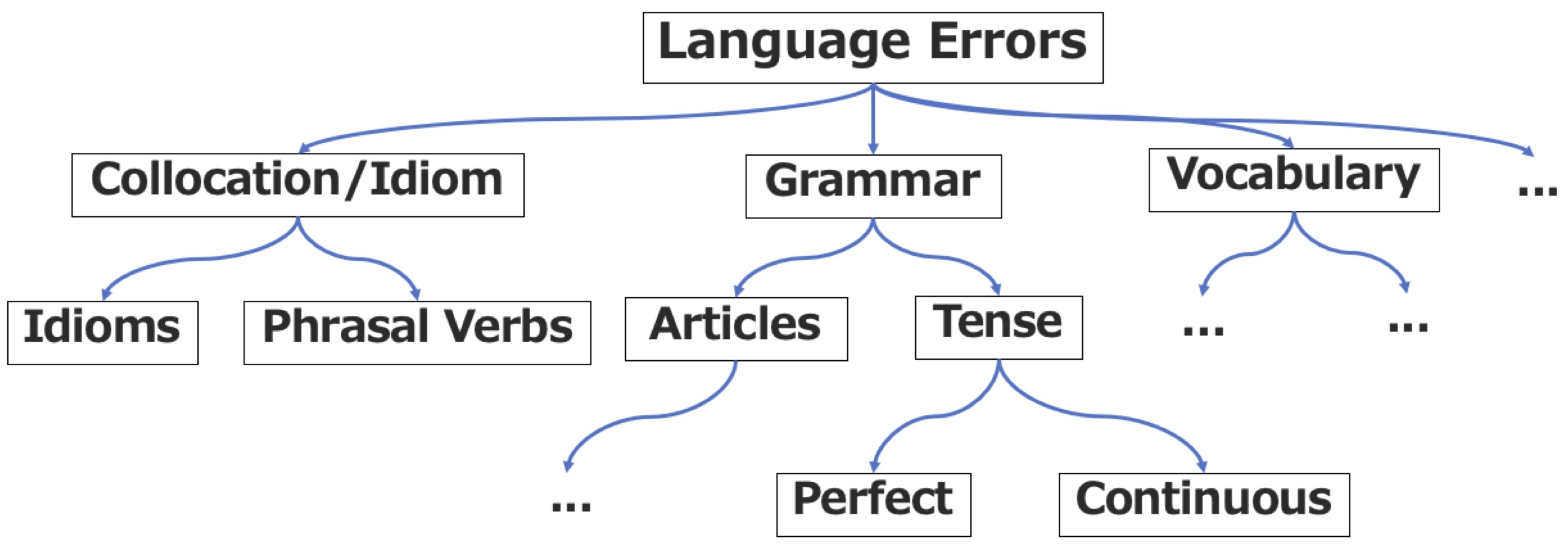}
    \caption{Sample of the our hierarchical learner error typology.
    Here, ``Idioms'', ``Phrasal Verbs,'' ``Perfect'' and ``Continuous'' serve as terminal tags.}
    \label{fig:typology_sample}
\end{figure}

\paragraph{Error Type} After examining previous error typologies (see Section \ref{sec:annotation_related_work}), we were motivated to define a new typology of learner errors that better reflects the \textit{topic} of WCF likely to be given in response to the error.
That is, the typology targets the perceived language knowledge gap underlying a given error.
A sample of the typology can be seen in Figure \ref{fig:typology_sample}, and the full typology can be found in the Appendix.
Errors are divided into six collections such as ``Vocabulary'' and ``Grammar,'' each containing sets of tags and sub-tags.
The hierarchical nature of the typology allows three things: (a) to group similar errors; (b) to analyze the output (e.g., classification success rates for each collection or downstream analysis of learner revision by error type); and (c) to toggle collections or individual tags to provide focused feedback.
In total, there are 81 terminal tags.

When naming tags, we prioritized alignment with concepts familiar to teachers of English as an additional language, such as terms commonly in textbook indices.
In fact, as these tags represent relevant educational concepts, it is possible to map them to such resources.
Thus, in our proposed feedback, shown in Figure \ref{fig:differentiated-feedback}, we include resource links, as our framework indirectly supports them.

\paragraph{Error Generalizability} 

Annotators judge whether an error represents a relatively predictable rule of the language, or is more idiosyncratic, perhaps specific to one lexical item, following the principles outlined in Section \ref{sec:strategies_for_feedback}.

\subsubsection{Feedback Annotations}

In addition to describing the learner's error, annotators provide WCF in response to it.
This process includes defining the comment location, writing the feedback comment, and assessing its directness.

\paragraph{Comment Highlight} 
As with previous works \cite{nagata-2019-toward,nagata-etal-2020-creating,nagata-etal-2021-shared,song-etal-2024-gee,kaneko-okazaki-2024-controlled-generation}, we annotate text spans to identify error locations, or more accurately, the scope of a feedback comment.
A comment highlight must contain all words to be edited, but may extend further to provide appropriate context.
For example, in ``We put *down* the fire,'' the minimal highlight would be ``down,'' but our guidelines suggest highlighting ``put down'' to complete the lexical unit of the phrasal verb.
Since the feedback is likely to discuss the two highlighted words as a set, our comment highlight includes both, adapting to the error in question.

\paragraph{Explanation and Edit Suggestion} The feedback is divided into two parts: the explanation ``identifies \textbf{what is wrong} and \textbf{why},'' while the edit suggestion ``tells the learner \textbf{what to do} to fix the issue.''
This separation enables targeted handling of either field. 
Otherwise, the default assumption is that the WCF consists of the explanation followed immediately by the edit suggestion.

\paragraph{Directness} Annotators judge whether their edit suggestion provides a direct correction or a hint.
We define a direct correction as one that provides the exact text of the suggested edit (e.g., ``Change `eat' to `ate''') or specifies the exact word(s) and location of an insertion or deletion, while a hint provides a metalinguistic clue (e.g., ``Change `eat' to the past tense'').

\subsection{Preliminary Dataset Collection}\label{data_collection}

To confirm the robustness of our framework and the quality of our annotation guidelines, we conducted a pilot annotation study.
Two annotators with English teaching experience annotated a subset of data in parallel in three batches.
Both of them were co-authors of this work, one of them being the first author.

The base data consisted of examples taken from the training set of a corpus of learner writing and corrections, EXPECT \cite{fei-etal-2023-enhancing}.
The instances were randomly selected as a representative sample across learner levels A1-C2 present in the dataset.
Annotation was performed in three batches of 114, 114, and 228 instances, for a total of 456 annotated instances.
The annotators discussed disagreements and updated the guidelines between the batches.

\subsubsection{Inter-Annotator Agreement}

We measure agreement for categorical annotations such as error type, generalizability, and directness with exact match rate and Krippendorff's Alpha \cite{krippendorff2019content}.
In the case of error type, we compare the terminal tag, the most fine-grained tag used for a given example.

For the comment highlights, we report exact matches and partial matches.
Partial matches are evaluated with pairwise token-level F1 scores.

For metrics other than exact match rate, instances rejected by both annotators are not included in the calculations, including the denominators of averages.
Instances rejected by one annotator but annotated by the other are considered mismatches (i.e., null vs. not null), and the instance is assigned a score of 0.

\begin{table}[t]
    \centering
    \caption{Inter-annotator agreement observed during the pilot annotation study.}
    {\fontsize{8pt}{10pt}\selectfont
    {\begin{tabular}{l@{\hspace{10pt}}c@{\hspace{10pt}}c@{\hspace{10pt}}c@{\hspace{10pt}}c}

        \toprule
        Annotation & Agreement Metric & Batch 1 & Batch 2 & Batch 3\\
        \midrule
        Error Tag & Exact Match & 63.16\% & 69.30\% & 76.32\%\\
        Error Tag & Krippendorff's $\alpha$ & 0.601 & 0.677 & 0.794\\
        Comment Highlight & Exact Match & 18.42\% & 51.75\% & 54.25\%\\
        Comment Highlight & Pairwise Token F1 & 0.375 & 0.699 & 0.778\\
        Generalizability & Exact Match & 70.18\% & 74.56\% & 80.26\%\\
        Directness & Exact Match & 62.28\% & 70.18\% & 80.26\%\\
        Rejections & Krippendorff's $\alpha$ & 0.366 & 0.541 & 0.645\\
        \bottomrule
    \end{tabular}}
    }
    \label{tab:annotation_results}
\end{table}

Table \ref{tab:annotation_results} shows the agreement across each batch.
Scores on all metrics improved between the batches, suggesting that the discussions and guideline updates were productive.
The most notable increase is seen in the comment highlight agreement, which increased significantly following a major rewrite of the relevant section of the annotation guidelines after the first batch.

The encouraging inter-annotator agreement achieved in the pilot study suggests that our annotation scheme is well-defined and can be effectively applied by human annotators. 
We are therefore confident in moving forward with annotating additional data as part of future work.

\section{Experiments}

To empirically validate the usefulness of this framework for feedback generation, we explore methods to incorporate our annotations into automated WCF pipelines.
The details of each approach can be seen in Section \ref{sec:system_experiments}.
Human raters evaluated the feedback from each, providing perspectives on its pedagogical validity.
The human rating experiment is discussed in Section \ref{sec:rating_experiments}.

\subsection{Feedback Generation}\label{sec:system_experiments}

\begin{figure}[!ht]
    \centering
    \begin{tcolorbox}[colframe=black!70, colback=gray!10, boxrule=1pt, width=\textwidth, left=2pt,right=2pt,top=2pt,bottom=2pt]
        \ttfamily \fontsize{6pt}{6pt}\selectfont
\#\#\# Example 1\\
\\
Example 1 Input:\\
source: The responsibility of <*the* educational institutions> is to make sure that he/she won't be in danger in the swimming pool instead of dissuading him/her from getting close to the water.\\
corrected: The responsibility of *[NONE]* educational institutions is to make sure that he/she won't be in danger in the swimming pool instead of dissuading him/her from getting close to the water.\\
error\_tag: \{One of: "Missing/Unnecessary Article" (Ours), or "Article" (EXPECT), or "U:DET (Determiner is not needed)" (ERRANT)\}\\
\\
Example 1 Output:\\
feedback\_explanation: The article "the" is not necessary because you are talking about all educational institutions in general.\\
feedback\_suggestion: Remove the article "the."\\
\\
\#\#\# Example 2\\
\\
Example 2 Input:\\
source: That is why I totally agree with Richardson's modality dealing with this important issue which is present in students', schools' and <parents *[NONE]* lives> nowadays.\\
corrected: That is why I totally agree with Richardson's modality dealing with this important issue which is present in students', schools' and parents *'* lives nowadays.\\
error\_tag: \{One of: "Possessive" (Ours), or "Possessive" (EXPECT), or "M:NOUN:POSS (Noun possessive is missing)" (ERRANT)\}\\
\\
Example 2 Output:\\
feedback\_explanation: When something belongs to someone, it is necessary to use a possessive.\\
feedback\_suggestion: Change "parents" to a possessive form to show whose lives we are talking about.
    \end{tcolorbox}
    \caption{The two static few-shot examples as they appear in our prompt.
    Additional examples follow the same format.
    \texttt{error\_tag} is omitted from the keyword-free setting.}
    \label{fig:few-shot_examples}
\end{figure}

We compare three approaches, all of which incorporate a large language model (LLM) for text generation: keyword-guided generation using error tags from our typology and existing typologies, template-based generation using fillable templates grouped by tag, and a baseline of keyword-free feedback generation.
All systems used the same underlying LLM, \texttt{gpt-4o-2024-11-20} \cite{openai2024gpt4ocard}, with temperature 0 and a response format of \texttt{"json\_object"}.
As input, all systems received the learner's original sentence with the error and correction marked with asterisks and the comment highlight marked with angle brackets.
Systems that consider error type tags received the input instance's tag as well.
As output, systems return separate \texttt{feedback\_explanation} and \texttt{feedback\_suggestion} fields.

For this experiment, we isolate the feedback writing task and set all prerequisite ``upstream'' information, such as successful error detection, classification, and highlighting, to the ground truth ``oracle'' data from the human annotations.
In a real-world setting, such information must first be obtained from the learner's text.
However, by keeping this information stable across all the feedback sources we compare, we can better focus on the characteristics of the feedback content itself, without attributing any issues to a prior stage.

The 228 instances comprising the first two annotation batches are designated as a ``Training'' set.
The other half, the 228 instances from batch 3, are designated as the ``Test'' set.
To avoid potentially flawed data, we exclude all instances rejected by either annotator, leaving the Training set with 189 instances and the Test set with 197.
Then, for each instance, we randomly selected one of the two annotations, leaving 50\% representation of each human feedback writer.

\subsubsection{Keyword-guided Systems}

We test a paradigm of keyword-guided feedback generation, in which the model is prompted to consider an error type label when writing feedback.
We define three such systems by providing tags from three different typologies: our typology outlined in Section \ref{sec:annotation_scheme}, the tags from the original EXPECT dataset (available from our base data), and ERRANT tags obtained by comparing the source and corrected sentences for each instance.

The system is provided with up to four few-shot examples of inputs and corresponding feedback comments, which are drawn from the Training set.
Two of the examples are static: one with an article error and one with possessive error.
These examples were chosen because all three typologies had clear tags for the errors in question.
They can be seen in Figure \ref{fig:few-shot_examples}.

Up to two other examples could be provided as well.
These were randomly selected from the instances that share the error type of the input instance, using the tags from the typology being evaluated.
When there are no annotated instances that share a tag with the input, the two static examples remain as a general demonstration of the task.

Some additional adaptations were made for the ERRANT setting.
In a few cases, ERRANT assigned multiple tags to an instance in the data.
This occurred in 11 of the 386 non-rejected instances used in the experiment, a rate of 2.84\%.
In these cases, few-shot examples were selected from the set of all instances that share an ERRANT tag with the input, favoring one example each from two different tags if possible.
Additionally, since ERRANT tags are expressed in codes such as \texttt{M:NOUN:POSS}, we append a short natural language description after each tag used in the prompt.
For these, we use the ERRANT tag descriptions from Li and Lan \cite{li-lan-2025-large}, and add our own for the \texttt{OTHER} tags, which they omit.

\subsubsection{Baseline Keyword-Free System}

The keyword-free system uses a prompt that is nearly identical to the keyword-guided systems, minus all mention of error types.
This system uses the same two static few-shot examples as the keyword-guided system, plus two more examples drawn randomly from the Training set.
Since error tags were not considered in the few-shot selection process, this system always received a total of four examples.

\subsubsection{Template-guided System}\label{sec:templates}

Works such as Lai and Chang \cite{lai-chang-2019-tellmewhy} and Coyne \cite{coyne-2023-template} describe using templates to control feedback generation.
A template is selected and any slots are filled by lookup mechanisms or neural models.
Potentially, this can mitigate issues described in older works in feedback generation, such as Hanawa et al.~\cite{hanawa-etal-2021-exploring}, who report generative outputs that mix multiple training examples or fail to adapt to the specific words in the input sentence.
To explore such an approach, we include a template-guided system that uses our error types.

\begin{figure*}[!t]
    \centering\includegraphics[width=0.9\textwidth]{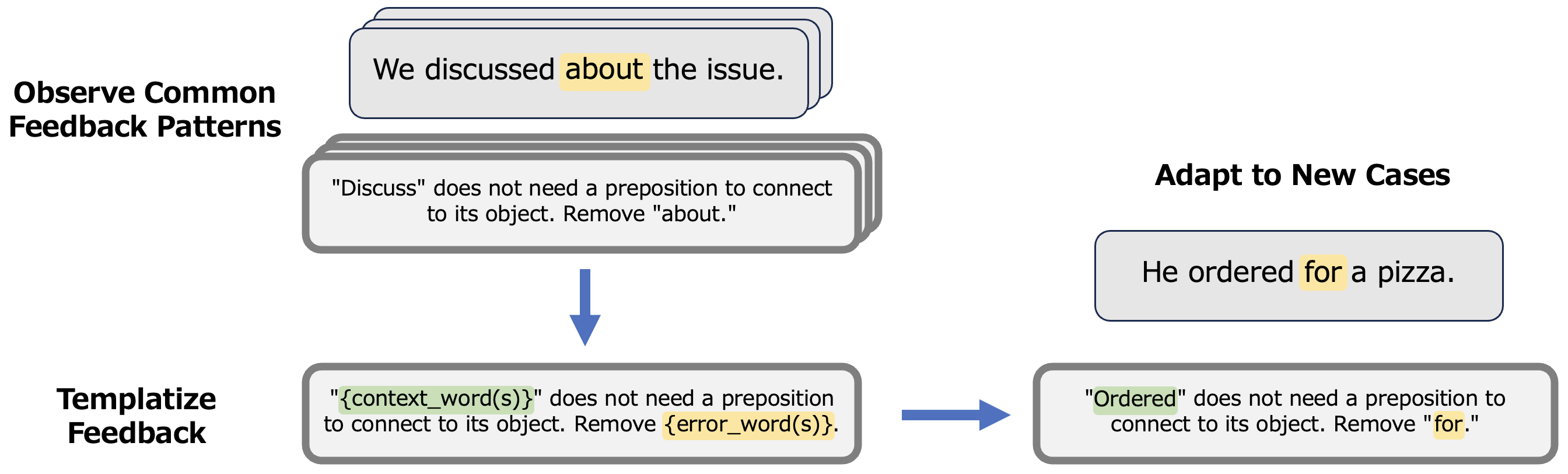}
    \caption{Visualization of the template construction process.
    Templates are written by hand based on patterns in our data.
    Here, \texttt{\{error\_word(s)\}} represent what must be removed or edited, and \texttt{\{context\_word(s)\}} are used to justify explanations.}
    \label{fig:templatization}
\end{figure*}

We create feedback templates by manually grouping existing feedback comments by error type tag, observing common patterns, and writing a template based on each ``archetype'' found, as depicted in Figure \ref{fig:templatization}.
Each archetype represents a set of instances under the umbrella of a given error tag that are similar enough that a template could substitute for any of them.
An error tag is assumed to have several such archetypes and, thus, several potential templates.

We start by converting the examples in our annotation guidelines into templates.
We then examine the Training set, adding any previously unseen archetypes.
This resulted 149 templates, 120 from the guidelines and 29 from the data.
We do not repeat this for the Test set, in order to keep it ``unseen.''
This process achieved 92.63\% coverage of the Training set (remaining errors were judged to be rare and specific, so no template was composed), and 76.76\% of the Test set.

For the purpose of the template system experiment, each instance was assigned a ground-truth correct template.
At inference time, the system is shown all templates associated with the input instance's error tag and asked to select the most appropriate option.
Concretely, the system outputs a \texttt{template\_id}, and then writes the feedback fields by filling the template.
Instances with no template coverage were assigned a \texttt{"None"} template, and the system was expected to select this option and refrain from giving feedback in these cases.

\subsection{Human Evaluation}\label{sec:rating_experiments}

To assess the feedback for pedagogical validity, we arranged a task in which annotators rated feedback comments on several criteria, as seen in Table \ref{tab:comment_criteria}.
We recruited four raters, who each had at least 7 years of experience teaching English as an additional language, including writing instruction in high school and university settings.
The raters were not co-authors of this paper, but belong to the same university as the first author.
They were not provided with the source of each feedback comment or otherwise directed to favor ``our'' systems.
The raters were paid an average of \$19.31 USD per hour of work performed.

\begin{table}[!t]
    \centering
    {\fontsize{8pt}{10pt}\selectfont
    \caption{Feedback Rating Criteria. ``Out of scope'' content includes e.g., assumptions about the learner or AI chat-like phrases such as ``Okay, here's my feedback:''}
    \begin{tabular}{lc}
        \toprule
        \textbf{Rating Criterion} & \textbf{Annotation Type} \\
        \midrule
        Comment is relevant to the error & Binary Radio \\
        Comment is factually correct & Binary Radio \\
        Comment explains what is wrong and why & Binary Radio \\
        Comment explains what to do to fix the error & Binary Radio \\
        Comment is comprehensible to a CEFR B1-B2 academic learner & Binary Radio \\
        Comment contains ``out of scope'' content & Binary Radio \\
        Whether the comment is direct, a hint, or NA & Radio (3 options)\\
        Overall quality of the comment & Likert (1-5) \\
        Rater Comments & Text Box \\
        \bottomrule
    \end{tabular}
    }
    \label{tab:comment_criteria}
\end{table}

The raters were provided with a document detailing the task and each rating field, which is available in the Appendix.
They attended a paid training session in which the first author demonstrated the task and interface, answered questions, and monitored ratings on practice instances.
Annotation was then performed in three batches with quality reviews based on a sample of 10\% of the batch plus the set of a) all instances with a rater comment, b) all instances where human-written feedback was rated as low-quality, and c) all instances where the rater disagreed with the reference directness label for human or template feedback.

\subsection{Results}\label{sec:results}

\subsubsection{Feedback Generation Results} 

The template-guided system demonstrated mixed performance in template selection and generation tasks.
While it correctly identified the ground truth template in 76.65\% of cases, yielding an F1-score of 0.76, it struggled to handle cases where no template was appropriate.
When the correct action was to refrain from providing feedback (selecting the \texttt{"None"} template), the system often incorrectly attempted to fill an available template, resulting in low recall (0.37). 
While the system did correctly identify some cases requiring no feedback (all \texttt{"None"} outputs were correct), this performance suggests limited robustness in identifying coverage gaps and withholding feedback.

Additionally, we observed template filling errors (e.g., including template syntax such as curly braces in the final text) in 4.57\% of outputs.
While these outputs can be filtered out programmatically before reaching learners, they represent an additional failure mode that must be taken into consideration.

\subsubsection{Human Evaluation Results} 

\begin{table}[t]
    \centering
    {\fontsize{8pt}{10pt}\selectfont
    \caption{Average ratings for each feedback source. Most values are relatively close.}
    \begin{tabular}{lc@{\hspace{6pt}}c@{\hspace{6pt}}c@{\hspace{6pt}}c@{\hspace{6pt}}c@{\hspace{6pt}}c@{\hspace{6pt}}c}
    \toprule
     & Relevant & Factual & What \& Why & What to Do & Comp.& Scope ↓ & Overall \\
    \midrule
    Human & \textbf{1.000} & 0.972 & 0.987 & \textbf{1.000} & 0.952 & 0.008 & 4.449 \\
    Our Tags & \textbf{1.000} & 0.970 & 0.992 & \textbf{1.000} & 0.970 & 0.008 & 4.487 \\
    ERRANT & 0.997 & 0.967 & 0.992 & \textbf{1.000} & \textbf{0.982} & \textbf{0.003} & 4.475 \\
    EXPECT & 0.997 & \textbf{0.975} & 0.990 & \textbf{1.000} & 0.975 & 0.005 & \textbf{4.500} \\
    No Tag & 0.995 & 0.970 & \textbf{0.997} & \textbf{1.000} & \textbf{0.982} & 0.005 & 4.495 \\
    Templates & 0.977 & 0.921 & 0.944 & 0.994 & 0.980 & 0.023 & 4.184 \\
    \bottomrule
    \end{tabular}
    }
    \label{tab:mean_ratings}
\end{table}

The results of the human rating task can be seen in Table \ref{tab:mean_ratings}.
All systems were rated fairly well in this task, with mean quality ratings between 4 and 5 points.
No output from any system was reported as toxic or inappropriate.
Keyword-guided systems and the keyword-free system performed similarly, and the specific typology used was not a significant factor.
Interestingly, they were rated comparably to the human feedback writers.

\begin{figure*}[t]
    \centering
    \includegraphics[width=\textwidth]{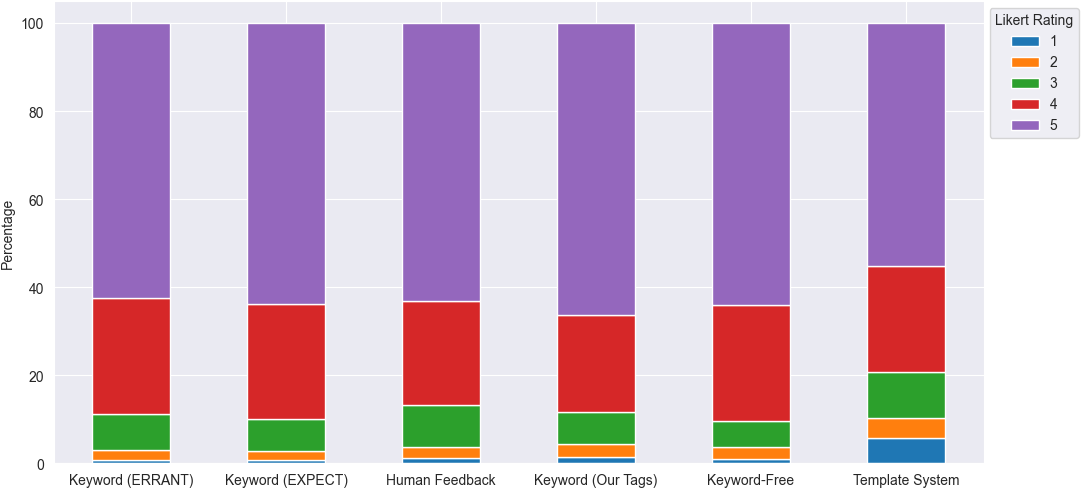}
    \caption{Distribution of overall quality ratings for different feedback sources.}
    \label{fig:feedback_rating_distribution}
\end{figure*}

Figure \ref{fig:feedback_rating_distribution} shows the breakdown of quality ratings by source.
The template system has the highest proportion of feedback comments rated 1 or 2.
Examining such cases, the most common pattern is that there were no completely valid templates for the instance, but the system attempted to fill one instead of selecting \texttt{"None"} and refraining from writing feedback.
The average quality rating is 4.40 when the template is chosen correctly, 3.51 when chosen incorrectly, and 2.88 when a template is chosen and filled when \texttt{"None"} is the correct choice.

Despite some language in the prompt to encourage hint feedback, and the presence of in-context examples providing hints, the keyword-guided and keyword-free systems almost always provided direct corrections.
The human feedback writers provided hints for 40.86\% of instances, but these systems ranged from 0 to 3\%.
The template-guided system was closer to human practices in this regard, giving hints in 39.77\% of instances, for an F1 score of 0.78 for hint feedback.

\subsection{Discussion}

All models received high ratings, suggesting that LLMs, when properly constrained, can generate valid feedback in many cases.
However, even the best-performing system did not achieve 100\% factuality.
It is imperative to exercise caution generating ``pedagogical'' text and promote students' AI literacy skills.

For most settings, the feedback was overwhelmingly direct.
This did not preclude it from being rated well, but does suggest an area where the AI and human feedback writers differed significantly on qualitative grounds.

The keyword-guided and keyword-free results suggest that the typology used was not a significant factor in feedback quality in this experiment.
Further investigation is needed to distinguish whether this is due to a similar informativeness level of all typologies, or a bias or keyword insensitivity in the base model.

The quality issues resulting from template coverage gaps present a challenge for such systems.
Nevertheless, our template system was effective for controlling directness, and thus shows some promise for contexts where this is a priority.

\section{Conclusion}

We define a framework to annotate data with key factors of errors and feedback and explore their impact on LLM-based automated feedback.
We find that human teachers rate all systems fairly well, that error tags had little effect, and that a template-guided approach reflected human directness decisions but faced challenges handling template coverage gaps.
Limitations include a lack of student ratings or learning outcomes and the use of oracle information for error location and classification.
In future work, we will investigate student responses to feedback systems, methods to control directness without 
sacrificing generation quality, and performance in fully automated settings
requiring error detection.

\subsubsection*{Appendix} Available at \url{github.com/coynestevencharles/annotating-errors-wcf}

\begin{credits}
\subsubsection{\ackname} This work was supported by JSPS KAKENHI Grant Numbers JP22H00524 and JP25K03175.

\subsubsection{\discintname}
The authors have no competing interests to declare that are relevant to the content of this article.
\end{credits}

\bibliographystyle{splncs04}
\bibliography{bibliography}

\end{document}